\documentclass[onecolumn,11pt]{article}

\usepackage{natbib}
\usepackage{enumitem}

\usepackage{microtype}

\usepackage{times}
\usepackage{tabu}
\usepackage{courier}
\usepackage{soul,color}

\usepackage{hyperref}
\usepackage{url}
\usepackage{breakurl}

\usepackage{bussproofs}

\usepackage{color}
\usepackage{graphicx}

\usepackage{caption}
\usepackage{subcaption}

\usepackage{amsmath}
\usepackage{amsthm}
\usepackage{amssymb}

\usepackage{xypic}

\usepackage[all,2cell]{xy}\UseAllTwocells\SilentMatrices
\usepackage{multirow}

\theoremstyle{plain}
\newtheorem{theorem}{Theorem}

\newtheorem{proposition}[theorem]{Proposition}

\theoremstyle{definition}
\newtheorem{example}{Example}

\newtheorem{definition}{Definition}

\def\shf{\mathcal}

\begin{document}

\title{\textbf{A Sheaf Model of Contradictions and Disagreements. \\
Preliminary Report and Discussion. \\
}}

\author{ Wlodek Zadrozny$^{1}$, Luciana Garbayo$^{2}$ \\
$^1$Department of Computer Science, UNC Charlotte \\
$^2$Departments of Philosophy  \& Medical Education, U. of Central Florida \\
Corresponding authors: wzadrozn@uncc.edu, Luciana.Garbayo@ucf.edu
}


{}

\maketitle

\begin{abstract}
We introduce a new formal model -- based on the mathematical construct of sheaves -- for representing contradictory information in textual sources. This model has the advantage of letting us (a) identify the causes of the inconsistency; (b) measure how strong it is; (c) and do something about it, e.g. suggest ways to reconcile inconsistent advice. 
This model naturally represents the distinction between \emph{contradictions} and \emph{disagreements}. It is based on the idea of representing natural language sentences as formulas with parameters sitting on lattices, creating partial orders based on predicates shared by theories, and building sheaves on these partial orders with products of lattices as stalks. 
Degrees of disagreement are measured by the existence of global and local sections.

Limitations of the sheaf approach and connections to recent work in natural language processing, as well as the topics of 
contextuality in physics, data fusion, topological data analysis
and epistemology are also discussed.
\footnote{This paper was presented at ISAIM 2018, International Symposium on Artificial Intelligence and Mathematics. Fort Lauderdale, FL. January 3–5, 2018. Minor typographical errors have been corrected. The authors retain the copyright of this work.}

\end{abstract}

\setcounter{secnumdepth}{2}

\section{Introduction: Modeling disagreements}

\color{black} 

\subsection{Motivation}
The motivation for this and related paper \cite{ZadroznyIWCS2017} comes from our need to model the contents of different and often contradictory documents pertaining to the same topic or decision. One, perhaps surprising, example of such situation is in representing the content of medical guidelines, where guidelines created by different accredited medical societies often contradict each other. 
These disagreements raise uncertainty in disease screening and treatment; the lack of guidelines consistency is confusing for patients and doctor, and also contributes to overdiagnosis and overtreatment. This conundrum for one particular domain is neatly  summarized in a CDC table comparing "Breast Cancer Screening Guidelines for Women", where guidelines from seven different accredited medical bodies are presented.  
\footnote{\url{https://www.cdc.gov/cancer/breast/pdf/BreastCancer ScreeningGuidelines.pdf}  -- retrieved 10/7/2017.} 
From this table we get our first example:  
\begin{example}
Contradictory recommendations for \emph{"women aged 50 to 74 with average risk"} coming from three (of the seven) different organizations:
\begin{itemize}
\item[(a)]\emph{Screening with mammography and clinical breast exam annually.} 
\item[(b)]\emph{Biennial screening mammography is recommended.}
\item[(c)]\emph{Women aged 50 to 54 years should get mammograms every year.
Women aged 55 years and older should switch to mammograms every 2 years, or have the choice to continue yearly screening.}
\end{itemize}
\end{example}

In addition to early detection guidelines, there are many cases where starting a therapy or other action is based on contradictory criteria. For instance, until recently 
WHO guidelines for HIV management 
recommended starting antiretroviral therapy if the CD4+ T-cell count is below 350/mm³. The 
American guidelines have recommended starting antiretroviral therapy even when the CD4+ T-cell count was above 500/mm³, and for anyone infected with HIV. WHO recently moved their guidelines closer to the American version.\footnote{\url{http://apps.who.int/iris/bitstream/10665/204347/1/WHO_HIV_2015.44_eng.pdf?ua=1} retrieved 10/7/2017} \\

\begin{sloppypar}
Almost any set of recommendations will have inconsistencies, often reflecting expert positions and different knowledge bases. Even for a common task like cooking bacon, there could be disagreements about the proper time and temperature of the oven.
\footnote{\url{http://lifehacker.com/the-best-way-to-make-flat-crispy-bacon-for-sandwiches-1788504821} \url{http://lifehacker.com/5711834/ditch-the-skillet-fire-up-your-oven-to-cook-perfect-bacon}  -- last retrieved 10/7/2017 }


Similar problems appear in many aspects of natural language processing. For example, written accounts of events often disagree about important details, making multi-document summarization challenging. Parsing of sentences and the translation of the parse into a logical form often produces a correct formula that does not exactly match other formally represented knowledge, even when this knowledge is coming from well engineered sources, such as Wordnet. For longer sentences, very frequently, only parts of the formal interpretations are correct. This is due to the fact that language use, e.g. word choice, is highly contextual, and many different structures can be associated with the same string of words.\\

In all of these cases simply saying that there is an inconsistency in conveyed information feels unsatisfying. Thus, we would like to (a) be able to say what causes the inconsistency; (b) measure how strong it is; (c) do something about it, e.g. suggest ways to reconcile inconsistent advice. 

\end{sloppypar}

\subsection{Introducing the distinction between contradictions and disagreements}

The natural language processing community has so far focused on finding explicit contradictions in texts, e.g. \cite{de2008finding}
\cite{williams2017broad}. However, we need a finer tool -- one capable of modeling the degree of contradiction. This is why we introduced the distinction between contradictions and disagreements, and modeled it using a lattice-based extension of propositional logic  \cite{ZadroznyIWCS2017}. The model introduced there can account for the intuition that while the medical guidelines in Example 1 (a) and (b) are definitely contradictory, the ones in Example 2 below can be reconciled, e.g. by making sure that the person follows the more demanding guidelines. 

\begin{example}
\label{ex:exercise}

Exercise recommendations for adults often differ in details: 
\footnote{\url{https://www.supertracker.usda.gov/physicalActivityInfo.aspx}. \url{https://www.cdc.gov/healthyweight/physical_activity/}
\url{http://www.mayoclinic.org/healthy-lifestyle/fitness/expert-answers/exercise/faq-20057916} -- last retrieved Oct 7 2017 }
One organization may recommend a minimum of 150 minutes per week, another 150-300  minutes per week, and yet another minimum 30 minutes per day (we simplify the recommendation a bit here).

Clearly, someone exercising 30 min per day (or a bit more), 7 days a week, likely satisfies all three guidelines. The guidelines don't agree 100\%, but intuitively they are not 100\% contradictory either. 

\end{example}

The main idea we would like to convey with this example is that many expressions can be modeled as \emph{ types with natural partial orders} \cite{ZadroznyIWCS2017}. This is true of temporal expressions, distances, intensities, number of people in the crowd, etc. Moreover, usually these expressions admit minimum ($\wedge $, least upper bound) \ and maximum functions, that is, form a \emph{lattice} (or at least a semi-lattice where $\wedge$ is defined). Therefore, from now on, we can restrict ourselves to talk formally about logical formulas with parameters of different types, with each type represented by a lattice. For example, "recommended daily exercise time is 20-30 minutes" can be represented as  $$exercise(Minutes:[20,30])$$

\color{black}
\subsection{Novel ideas in this paper}

In this paper we introduce an alternative, and perhaps better, formalization of the distinction between contradictions and disagreements. 
This new approach is based on sheaves. 
A sheaf is a mathematical and computational tool for systematically tracking locally defined data and information flow. In our case, 'locally defined' corresponds to information being given by different documents. Sheaf-based models can give us a principled, topological approach to account for multiple sources of information. They also permit clear modeling of data transformation, through the introduction of mappings between elements of a sheaf, represented by nodes of a graph (as we shall see below). Thus a sheaf-based model can be used to track and reconcile disagreements in data, explicitly describe mappings between data, as well as support approximate reasoning. 

However, sheaves very rarely appear as models of natural language phenomena. We are aware of two models: sheaves were used to model pronoun references in discourse \cite{abramsky2014semantic} and  presheaves \cite{FernandoScales2014} to model scales, e.g. 'fluents' -- descriptions of events involving time. 
However, sheaves do not appear in papers in the aclweb.org repository. 

We believe sheaves deserve more attention as a mathematical formalism to represent relationships between data elements produced by different sources, as is often the case in natural language processing.\\

In the context of data fusion, such work has been started by M.Robinson and others. In another context, sheaves have been used by S. Abramsky and others to represent quantum mechanical paradoxes, contextuality and related physical phenomena 
(\cite{abramsky2011cohomology}, \cite{abramsky2011sheaf}, \cite{caru2017cohomology}). In this paper we follow the approach of \cite{robinson2016sh}.\\

The use of sheaf-based model has the advantage of letting us (a) describe what causes the inconsistency; (b) measure how strong it is; (c) and do something about it, e.g. suggest ways to reconcile inconsistent advice. 
This model naturally represents the distinction between \emph{contradictions} and \emph{disagreements} based on the existence of global sections.  \\

\noindent\textbf{Plan of the paper:} We will keep the exposition only as formal as absolutely necessary, and will instead focus on the intuitions we associate with the sheaf-based model of contradictory theories. We introduce the sheaf formalism and a simple example in Section 2. Section 3 does the bulk of the formal work. Since the formalism is on the heavy side, it's worth discussing its pluses and minuses (Section 4). Section 5 contains a comparison with related work not discussed in Section 4, mostly in context of our ongoing work. We end with more details on ongoing work, a few questions, and conclusions (Sections 6 and 7).

\color{black}

\section{Formalism and Approach}

We will consider \textit{sheaves of finite theories, represented by lattices,} on partially ordered sets reflecting connections between theories through shared predicates. We would like to use the sheaf construction to find a maximal 'sensible' theory expressed in the language of parameterized logical theories, reflecting both the assertions of the original two theories, and the partial order, as provided by lattices of parameters. The concepts we will be using for this purpose are \textit{global section} and \textit{local section}. The intuitive idea is to create a set of restriction functions that would maximize the agreement. Later we will use the same idea to find the 'witnesses' to disagreements or contradictions. \\

We will start with definitions and examples, based on the exposition of \cite{robinson2016sh} and Robinson's  lectures, in particular {\url{http://www.drmichaelrobinson.net/sheaftutorial/}}, which focus on partially ordered sets (posets) -- this is the situation we are trying to model.

\begin{definition}
\label{df:sheaf_poset}
  Suppose that $P=(P,\le)$ is a poset.  A \emph{sheaf} $\shf{S}$ of sets on $P$ satisfies the following conditions:
\begin{enumerate}
\item For each $p \in P$, there is a set $\shf{S}(p)$, called the \emph{stalk at} $p$, 
\item For each pair $p \le q \in P$, there is a function 
 $$\shf{R}_{p \le q}:\shf{S}(p)\to\shf{S}(q)$$
 called a \emph{restriction function} (or just a \emph{restriction}), such that
\item For each triple $p \le q \le r \in P$,
$$\shf{R}_{p \le r} = \shf{R}_{q \le r} \circ \shf{R}_{p \le q}$$
\end{enumerate}

\begin{quote}
"When the stalks themselves have structure (they are vector spaces or topological spaces, for instance) one obtains a sheaf \emph{of} that type of object when the restrictions or extensions preserve that structure.  For example, a \emph{sheaf of vector spaces} has linear functions for each restriction, while a \emph{sheaf of topological spaces} has continuous functions for each restriction."[ \cite{robinson2016sh}]\\
\end{quote}

\end{definition}

Alternatively, we could express the idea of a sheaf as  the "existence of gluing" and the "uniqueness of gluing"; that is, we can glue together together restriction functions specified on any two open sets, if these restrictions agree on the intersection of the sets; and that we can do the gluing only in one way.\footnote{An intuitive exposition of sheaves can be found e.g. in \url{https://tlovering.files.wordpress.com/2011/04/sheaftheory.pdf}}. However, since in this paper we are only dealing with posets we can restrict our attention to the simple formal machinery defined above.

\subsection{Example sheaf and its sections}

This subsection tries to establish some intuitions about sheaves and restriction functions that will be useful later. 

\begin{example} 
\label{ex:ex123}
Consider an 
 order consisting of numbers $0, 1, 2, 3 $ with the usual "less than" relation: 

$$   \xymatrix{
    0   \ar[r]  & 1 \ar[r] & 2 \ar[r] & 3 \\ }      
$$

\noindent Let the stalk $\shf{S}(n)$ consist of the sets of either even or odd (but not both) numbers $\le n$. Then we have  
    $\shf{S}(3)= \{ \{1,3\}, \{1\}, \{3\}, \{0,2 \}, \{0 \} , \{2 \}\}$,
       $\shf{S}(2)= \{ \{1\}, \{0,2 \} ,\{0 \}, \{2 \}\}$,
              $\shf{S}(1)= \{ \{1\}, \{0 \}\}$,
                     $\shf{S}(0)= \{ \{0\}\}$.

Let each restriction function
$\shf{R}_{x \le y}:\shf{S}(x)\to\shf{S}(y)$ be a function that adds $ y-x $ to each number in an element of the stalk  $\shf{S}(x)$.
Then, for example,
$$ \shf{R}_{1 \le 3} ( \shf{S}(1) )= \{ \{3\}, \{2 \}\} \ \subset \ \shf{S}(3) $$


Note that  $\shf{R}_{x \le y}$'s are indeed restriction functions; that is, we have: if $p \le q \le r \in P$, then   $\shf{R}_{p \le r} = \shf{R}_{q \le r} \circ \shf{R}_{p \le q}$.

Also note that if we defined the restriction functions as e.g. adding $1$ (not $x-y$), when $x-y \neq 0$, the structure would not constitute a sheaf:  
$\shf{R}_{1 \le 3} \neq \shf{R}_{1 \le 2} \circ \shf{R}_{2 \le 3}$
because the composition of the two functions increments elements by $2$, while the $\shf{R}_{1 \le 3} $ increments them by $1$.

\end{example}

\color{black}

\begin{definition}
\label{df:gsection}
A \emph{global section} of a sheaf $\shf{S}$ on a poset $P$ is an element $s$ of the direct product $\prod_{x \in P}\shf{S}(x)$ such that for all $x \le y \in P$ then $\shf{R}_{x \le y}\left(s(x)\right) = s(y)$. 
A \emph{local section} is defined similarly, but is defined only on a subset $Q \subseteq P$.

\end{definition}

\begin{example}
Consider Example \ref{ex:ex123} again. Can we produce a global section? Yes, starting with $\shf{S}(0)$, the restriction functions uniquely determine the only global section consisting of the sequence
 $ ( \{0 \} , \{1 \}\  ,  \{2\}, \{3\} \ )$. 

Clearly there are also multiple local sections. 
However, as we shall shortly see looking at models of contradictory theories, not every collection of restriction functions produces a global section.

\end{example}

\color{black}

\subsection{Simple sheaves representing theories}

Above, we have seen that elements of a partial order can be associated with a set, and we can define pretty much arbitrary mappings on these sets, provided they can be composed properly to form restriction functions. 
Now, we would like to use sheaves to talk about contradictions and disagreements in theories.

\begin{example}
\label{ex:2th}
In the simplest non-trivial case we have two theories, $o=\{p(a)\}$ and $o'=\{p(b)\}$, each consisting of a single proposition, for example $p$ might be $$exercise(Minutes: \ \ )$$
Since the values of the parameters, in this case of the type $Minutes$, interpreted as intervals of natural numbers, form a lattice, we can compute their minimum, as the intersection of the intervals. Then if  $ a \wedge b \ne \bot$ we would like to produce $p(a \wedge b)$ as the 'region' of agreement. Note that, intuitively, we need to take advantage of the information that $a$ is associated with one theory and $b$ with another. Our sheaves and restriction functions need to reflect this fact.\\

So let's define a very simple sheaf. The domain we are considering consists of two points $o$ and $o'$, and the set $\{o, o'\}$. 
The ordering is given by $o, o' \le \{o, o'\}$; the connection $o - o'$  represented by $\{o, o'\}$ comes from the shared predicate $p$.

We attach to $o$ the partial order $\{x: x \le a \} $, and to $o'$ the partial order $\{x: x \le b \} $, representing the set of values compatible with the parameters of the given theory. That is, if the parameter $a$ says \emph{20-30 minutes}, the partial order will have all  intervals, expressed in minutes, in this range, arranged by inclusion. To $\{o, o'\}$ we attach the union of the two previously defined sets with their partial orders. This defines the stalks.

We need to define the restriction functions in a way that would give us a section capable of producing $p(a \wedge b)$.

We take the restriction function from $\shf{S}(o) $ to $\shf{S} (\{o, o'\})$ as the identity function.
This makes sense, since the domain of the former is included in the latter. We do the same for $o'$.  There are possibly many global sections, produced by identity restriction functions on elements $\le a \wedge b$. Since we have a natural induced partial order, we can choose the maximum such function, equal to $ a \wedge b$.

This approach allows us now to assert $p(a \land b)$.  On the other hand if $ a \wedge b = \bot$, the theory $\{ p(\bot) \}$ represents a genuine contradiction.\\ 
\end{example}

\textbf{Note}. If the types can be transformed, e.g. "average number of minutes per day" and  "average number of minutes per week" we can use restriction functions to translate the units e.g. "multiply by 7". The conversion would help with the statements from Example \ref{ex:exercise}, where we find reference to both minutes per week and minutes per day.

\color{black}

\section{Using sheaves to find agreements between theories}

We have seen how a simple sheaf on the lattice of parameters of formulas of two theories can be used to explicitly model a disagreement. 
Now we want to show how to create sheaf-based models in a more complex case. 

\subsection{Sharpening intuitions about sheaves of theories}

\begin{example}
\label{ex:3theories}
Let us return to Example 1. We translate the text of the screening guidelines into logical formulas using the following abbreviations: 
$s$ -- screening; $bi$ -- biennial; $an$ -- annual; $bx$ -- breast exam; $m$ -- mammography. Keeping the types of the arguments implicit e.g. representing $Age:[50-74]$ as simply $[50-74]$, we can represent the three sets of guidelines as three theories. 
Thus (a) 
\emph{Screening with mammography and clinical breast exam annually}  gets translated into $T_a$, and similarly (b) and (c) of Example 1.  

\begin{itemize}

\item[] $T_a =  \{ s([50,74],m,an), s([50,74],bx,an) \}$
\item[] $T_b = \{  s([50,74],m,bi) \} $
\item[] $T_c = \{  s([50,54],m, an) ,$ 
$s([55,74],m, bi)  \lor  s([55,74],m, an)  \} $\\
\end{itemize}

This situation is slightly more complex than before: we have multiple parameters, a disjunction and a split of the age interval in $T_c$. However, our analysis proceeds similarly as in Example \ref{ex:2th} . We start with the observation that $bi \wedge an = \bot$ and that $ bx \wedge m \neq \bot $, since the two exams can happen together. Also $ [55,74], [50,54]$ are both $ \le [50,74] $.

We will be attempting to build a sheaf on a partial order, and investigate the existence of sections: if a global section exists, the theories merely disagree, and if there is no global section, the theories are contradictory. Local sections show which theories can be reconciled.\footnote{The existence of unique global sections makes a presheaf into a sheaf. But we don't want to discuss these differences more formally in this expository presentation.}

The domain we are considering consists of all non-empty subsets of the set of three points $a$, $b$, and $c$, representing the three theories. 
As in Example \ref{ex:2th}, the ordering is by inclusion, e.g. $\{a\}, \{b\} \le \{a, b\} \le \{a, b\, c \} $.

We attach to each each point $e$ in the domain a subset $\shf{S}(e )$ of the product of lattices of the parameters, that is  of the set 
$P = Age \times \{m,bx, m \wedge bx\} \times \{bi, an\} \ $, where $Age$ is the set of age intervals ordered by inclusion. 
We set $\shf{S}({a} )$  to the values consistent with the parameters of the theory $T_a$,
\footnote{That is, as before, in Example \ref{ex:2th}, parameters of proper types that are $\le$ than parameters explicitly mentioned in  $T_a$}
and similarly for $\shf{S}({b} )$,  and $\shf{S}({c} )$. We attach the whole product $P$ to the other elements of the partial order.  This defines the stalks.

Regarding restriction functions, we can define them as identity functions, as we did before in Example \ref{ex:2th} .

Now we can ask the question about the existence of a global section.
We can see that a global section does not exist, because  $  an \wedge bi = \bot $, and therefore the mappings from $\{a\}$ and $\{b\} $ into $\{a , b\}$ cannot produce a local section. 
On the other hand, we have a local section given by the mappings from $\{a\}$ and $\{c\} $ into $\{a , c \}$, because the stalks for both $\{a\}$ and $\{c\} $ contain  $\{m \wedge bx\} $ and $[50,54], [55,74]$. This gives use two local sections:
$([50,54],m \wedge bx , an)  $ and $([55,74],m \wedge bx , an)  $. Similarly, we have a local section given by the mappings from $\{b\}$ and $\{c\} $ into $\{b , c \}$ based on $([55,74],m, bi)$. 
\end{example}

\color{black}

\subsection{Sheaves for representing sets of theories}

Using the intuition developed through Example \ref{ex:3theories} we can design a procedure to deal with multiple parameterized theories. We will start with the case of a finite set of 
ground\footnote{no free variables in any predicate} theories, and later discuss a more general case.

\subsubsection{Dealing with multiple disagreements: positive atomic theories}

Let us consider the general case of a finite set of positive 
ground  theories $ T_o = \{p_k(...) \}_{k<N}$, $ T_{o'} = \{p_k(...) \}_{k<N'}$, ..., where each predicate has a corresponding set of types, whose values for different theories are likely to differ.  
Thus, for every $p=p_k$, we can compare $T_o $ and $T_{o'}$  :
\begin{center}
$p(A_1\colon a_1,...,A_i\colon a_i)$ \   and \  $p(A_1\colon b_1,...,A_i\colon b_i)  $
\end{center}
\bigskip

We extend this approach to deal with multiple theories by generalizing the procedure developed in Example \ref{ex:3theories} . Namely, we define a partial order on a set of subsets of $ O = \{o, o', o'', ... \}$. As before, $\{o, o'\}$ belongs to the domain of the partial order if the corresponding theories $ T_{o}$ and $ T_{o'}$ share a predicate. The partial order is induced by inclusion on so defined subsets of $O$. 

As before, we attach to each $o$ the subset of the product of types $A_p = \{p\} \times A_1 \times ... \times A_i $  that is consistent with the theory 
$ T_o$, i.e. with the product of the lattices $A_1\wedge a_1,...,A_i \wedge a_i$ (where $\wedge$ is applied to each element of the type). Note that we need $p$ in $A_p$, because the same type or parameters might appear in different formulas (e.g. "take this pill daily and that one every other day"). [ We will ignore the $p$ if only one formula is involved.]

As before, we attach to the non-singleton elements of the partial order $O$ the union of the full products $A_p$.

Finally, the restrictions are defined as identity functions. And as before, we can look for local or global sections.\\

Using a triple induction: on the number of theories, the number of predicates, and the number of parameters, we can show that: 

\begin{proposition}
\label{prop:sheafcons}
A section  on the elements $\bar{o} =\{o, o', ...\}$, defined by the procedure above, is a collection of parameters consistent
\footnote{that is, if we add the corresponding formulas replacing the original parameters by the parameters of the section to each theory, we will still have a consistent theory.}
 with all the theories $ T_{o}$, for  $o \in \bar{o} $.  \\
\end{proposition}

\begin{proposition}
\label{prop:sheafinc}
If the theories $ T_{o}$, for  $o \in \bar{o} $ are inconsistent with 
each other,\footnote{i.e. their union is not consistent} then
there is no global section on 
$\bar{o} ,\{o \}, \{o' \}, ...$. \\

\end{proposition}

Positive ground atomic theories are our most important case. They shows the power of the approach in representing contradictions;  and the general case (below) is a natural generalization. In addition, most texts containing recommendations or news are positive. In many cases, we can create a new positive ground formula representing a sentence with a negation; e.g. "don't take aspirin" can be represented as $not\_take(Drug:Aspirin)$.

\subsubsection{Dealing with multiple disagreements: an example\\}

To deal with the general case, we need to address the negation operator. We start with  an example:

\begin{example}
\label{ex:negTh}
Consider two simple theories 

$\{ s([50,74],m) \}  $ and $ \{ \neg s([55,74],m) \} $

What is the intuitive meaning of this situation? Clearly, there is no way of reconciling 
the contradiction on the interval $[55,74]$.
But the second theory is agnostic about $[50,54]$. In principle we can have two models, one in which this is a full contradiction, and another one, with a possible agreement on the interval $[50,54]$.

To talk explicitly about what can go or cannot go with something else, using the language or restriction functions, we add the Boolean type to the set of parameters. This converts the two theories into

$ T_o = \{ s([50,74],m, T ) \}  $ and $ \{ T_{o'} = s([55,74],m, F) \} $, with the usual proviso that $T \wedge F = \bot$.\\

\textbf{Solution 1: Strict interpretation}

The partial order is defined as before: $\{o\}, \{o' \} \le \{o, o' \} $.  There is no global section, because only 
the elements $([dd,DD],m, F)$ with $dd \ge 55$ and $DD \le 74$ are stacked about $o'$.\\

\textbf{Solution 2: Permissive interpretation}

To get the 'agnostic' reading, we need extend the Boolean type to allow 'undefined': $T, F  \le U $.

As before, starting with a theory, and extending with additional Boolean value as above, we consider the product of types 
$$A \times Bool  = A_1 \times ... \times A_i \times \{T, F, U \} $$

As before, for the element $o$, we consider the subset of this product that is consistent with the one predicate theory
$ T_o$, i.e. the product of the lattices $A_o = {s} \times A_1\wedge a_1 \times ... \times A_i \wedge a_i \times Boolean \wedge T$. We do the same for $o'$ except that we'll have $Boolean \wedge F$ at the end of the expression. So, far this construction would give use the strict reading. 

To get the permissive, 'agnostic' reading, we also add to ${A_o}$ those $(s, a, U)$ for elements $a$ of $A$ which are not $A_o$. Similarly for $A_{o'}$. This corresponds to saying, "I don't know" for those sets of parameters for which the truth value is not explicitly defined. 

As in Example \ref{ex:exercise}, we extend all $(a,U)$'s with $(a,T)$ and $(a,F)$ -- this can't cause any trouble since the theory has no opinion on $a$, and  $T, F  \le U $,  and therefore $Bool$ behaves like other types discussed there. 

We now define the restriction functions again as identities. Then the 'dominant' global section for our example corresponds, as expected, to 
$ \{ s([50,54],m, T) \} $.  That's because this time we have $([50,54],m, U) $ $([50,54],m, T) $for $o'$ (from the extension above). Also, $ ([50,54],m, T) $  is an element associated with $o$, being a compatible with the theory $T_o$. Since $\{o, o' \} $ has all combinations of parameters, the only 'maximal' global section produces $ ([50,54],m, T) $, and other global sections are 
$(I,m,T)$ where $I \subset [50,54]$. (As in standard practice, we define a 'maximal' or 'dominant' function as one that is $ge$ any other function in the set under consideration, under the standard order induced by the product of the lattices).\\
\end{example}

\subsubsection{General case of multiple disagreements of finite propositional theories} 

We deal with the general case of finite consistent propositional theories as follows: For a theory  $ T$, consider the set of its minimal models, i.e. a minimal set of typed constants that can be put in the atomic formulas to make them ground and true, and the truth assignments to these sequences of parameters. These sets are finite, because the theory and the types are finite. Each such model is completely described by the ground atomic sentences satisfied in it.

Now we can replace any original theory $T$ by a set of ground theories defining its models (if the original theory has no negation or disjunction, this replacement doesn't change the theory). This replacement is justifiable, because we are simply making explicit the ambiguity of the original theory.

Having done it for all theories under consideration, we have reduced the general case to the one previously considered of positive atomic theories. A global section in a sheaf model based on these theories would again correspond to consistency and agreement under all interpretations, and local sections define the partial or local consistency. And again, we could prove the correctness of this algorithm by induction on the number of arguments, the number of predicates, and the number of theories.

We could once more prove a Proposition asserting the correctness of the algorithm using induction on the number of arguments, the number of predicates, and the number of theories as in Propositions 1 and 2.

\color{black}
\section{Discussion: Arguments for and against a sheaf approach in natural language understanding}

A sheaf is a mathematical and computational tool for systematically tracking locally defined data. It is  a principled approach to account for multiple sources of information. However, as noted earlier, sheaves rarely if ever appear as models of natural language phenomena. This paper, and the cited earlier work of \cite{FernandoScales2014} and \cite{abramsky2014semantic}, show that sheaf-based models can offer some insights about natural language phenomena. 

We see the promise of the approach in its ability to account for multiple sources of information -- this makes it well suited for multimodal semantics, e.g. combining text with image, gesture, etc. (cf. e.g. 
\cite{joslyn2014towards}). Similarly, this approach can model measures of coders' agreement in producing textual annotations. 
In addition sheaves allow us to identify the sources and strength of contradictions, and try to reconcile the disagreement in data when it is possible. Also, in NLP, there's often the need for mediating between e.g. units of measurement, ontologies, vocabularies etc. -- restriction functions could naturally serve this role, as we noted in passim above.

Although we haven't discussed it in this paper, sheaves can deal with measurement errors and probabilistic information. This can be done by introducing metrics on lattice of parameters (that is in addition to $\le$ we have distances) and defining the tolerances for restriction functions  \cite{robinson2017sheaves}. 

The cited work of Abramsky et al. shows
sheaves can model quantum phenomena. In NL semantics, some cognitive phenomena arguably are best modeled with quantum-like representation (e.g. negative probabilities), as shown in the work of \cite{busemeyer2012quantum} and \cite{aerts2009quantum}.
One place to see the potential of sheaf-based approach in NLP would be to try them on data sets that combine multiple sources of information such as the movies question answering test \cite{tapaswi2016movieqa}, with the dataset containing video clips, plots, subtitles, scripts, and videos' descriptions.

On a more speculative note, because of the connection of sheaves with computational topology, sheaf-based models perhaps could be used to explore persistence and other phenomena in natural language text, although, again, in this space very 
little has been published, the only example we are aware of being \cite{zhu2013persistent}. At the same time, topological data analysis is a very active area of research and applications (cf. e.g. 
 \cite{carls2009top}, \cite{bubenik2015stat}, and also ayasdi.com), and -- intuitively -- topology should have some bearing on models of text, given the cognitive makeup of humans, as exemplified e.g. in common metaphors involving space and time \cite{lakoff2008metaphors}.

On the other hand, the sheaf-based approach to NLP is mathematically heavy; for example is it worth the effort to introduce the formalism of restriction functions if later we are only using the identity mappings? These are not the most interesting functions, perhaps.

Besides being heavy on the formal side, the practical advantages of the approach are unclear. For example, we don't know whether -- when accounting for corpus based differences in interpretation -- it would provide better results than a Bayesian approach, i.e. would logic and topology add anything to statistics? However we note the observation of 
\cite{joslyn2015conglomeration} that "when applied to data sources arranged in a feedback loop, Bayesian updates
can converge to the wrong distribution!"

Neither it is clear how easy it would be to apply sheaves to large amounts of textual data. In a related domain of topological data analysis (TDA) it's been observed that 
"the time and space complexity of persistent homology algorithms is one of the main obstacles in applying TDA techniques to high dimensional problems"  
\cite{chazal2015subsampling}.
But textual data represented e.g. by term-document matrices is very high dimensional, 
and it's unclear whether sampling methods along the lines proposed there (ibid.)
would help. \\

Obviously, both the arguments for and against sheaves in NLP can be viewed as open problems. And this is the view of the authors.

\section{Comparisons with related work}

This paper shows that there might be promise in using sheaves to model the relationships between incompatible theories representing texts with potentially conflicting information. We showed how the language of restriction functions, and global and local sections, can be used to represent genuine contradictions and disagreements (which are possible to patch) in these theories. 
\subsection{Contradictions in Natural Language Processing}
Our original motivation in \cite{ZadroznyIWCS2017} came from modeling medical guidelines, but, as noted above, contradictory sets of documents are pervasive. In the last ten years the field of computational linguistics recognized this phenomenon, and initiated  ongoing computational research on reasoning with textual and contradictory information (\cite{de2008finding}, \cite{williams2017broad}). The motivation for a new formalization of compatibility comes partially from an opportunity to improve on the methods proposed in \cite{zadrozny1991semantics}. 

Recent and relavant work includes \cite{kalouli2017correcting} who observe the presence of contradiction asymmetries -- that is situations when two sentences shown in one order are viewed (by humans) as contradictory, while in the opposite order are not; such asymmetries arise when the contexts of the sentences are underspecified. For example the document 6819.txt has two sentences with incompatible judgments:

A = There is no man on a bicycle riding on the beach

B = A person is riding a bicycle in the sand beside the ocean

Entailment A to B = A\_entails\_B

Entailment B to A = B\_contradicts\_A

This opens the question whether sheaves can be used these contextual effects. The work of \cite{abramsky2015contextuality} suggests they can. The contradictory judgments could be perhaps explained by different defaults or external information brought to bear on the situation, and they could be represented as local sections with different starting points. 

\subsection{Contradictory knowledge bases}
We also want to mention -- to follow a referee suggestion -- a large amount of prior art on reasoning with contradictory knowledge bases in AI, 
as exemplified by \cite{grant2008analysing}, \cite{subrahmanian2007general}, and \cite{grant2016analysing}. 
 
The cited papers are concerned with sets of issues similar to ours: (a) identifying causes of inconsistencies; (b) measuring their strength; (c) and resolving them. 
The main difference is that we are focusing on the reconciling disagreements through the lattice of parameters: in our approach, the allowable parameters of logical formulas form lattices, and in the cited papers, they are atomic objects. These atoms could, in principle, be collected into sets, and endowed with additional lattice or poset structure. Then, the insights of the cited works could help us choose most appropriate local sections, if global sections do not exist. This could be done,
for example, using real valued measures proposed in \cite{grant2016analysing} or preferences on partially ordered theories, as in 
\cite{zadroznyCI1994}, or some combinations of the two. Another difference and the limitation of our work is that we do not explicitly deal with quantified formulas. Thus, the work of \cite{grant2008analysing} can be used as a guideline for extension of our approach to first order languages. 
We intend to investigate these relationships and discuss results in a future work.

\subsection{Logic and philosophy}
Clearly we are not the first ones to apply sheaves to formal theories. For example, 
\cite{goldblatt2014topoi} Ch.14 shows a formalization of 'local truth' using sheaves, and a representation of modalities. (This connection was also noted in \cite{maclane2012sheaves}). 
This is relevant, since modal truth is another intuition associated with contradictory theories. Namely, we can think of theories as describing possible worlds, where the statements shared by all theories correspond to necessity, and the other ones express mere possibilities. 
In \cite{ZadroznyIWCS2017} we showed a very elementary inference mechanism for reasoning with partially contradictory information.

At this point it is not clear to us the relationship of the simple lattice-based inference rule introduced there to the recent work of 
\cite{kishida16} who is formalizing reasoning with contradictory information withing a framework of sheaves and category theory.
To understand both these connections is a topic of our ongoing work.

In addition, this line of work has a promise in epistemology of disagreement \cite{christensen2004putting}, by representing global versus local epistemic appraisals of experts argumentation \cite{Garbayo2018}, and providing a clearer model of complex inconsistencies in epistemology applied to the special sciences. The use of such representations in decision theory, in particular, would help clarify aspects of multi-expert, multi-criteria decision-making in medicine \cite{garbayo2014}.

\subsection{Contextuality in computer science and physics}
The work on contextuality in computer science and physics exemplified by \cite{abramsky2015contextuality} and previously cited works.  
It shows the application of sheaves and cohomology to modeling of incompatible measurements and assignments of values to variables.
Cohomology is also discussed as a tool for sensor data fusion in the cited works of Robinson. These approaches are likely to also be applicable to language understanding, and, again, we are planning to look into cohomological models in a near future. 
\footnote{\cite{caru2017cohomology} gives a detailed exposition of using cohomology to find obstacles to global sections.}

\color{black}

\section{Ongoing Work and Open Questions}

\subsection{Modalities, cohomology and computation}
At this point, we are looking into possible uses of  modalities to  represent contradictory information. Intuitively, there is a connection, but it might interesting to see what modal models would emphasize. 

The second topic of our research is to clarify a possible role of cohomology in representation of contradictory textual information. Again, the intuitive and formal connections are there, where obstructions to global sections can be computed using cohomology (e.g. \cite{caru2017cohomology}, \cite{robinson2014topological}). Also, we have some dualities to explore. In this paper, we created the partial order on which the sheaves were build by linking theories that share a predicate. But, are there insights in another way of combining theories, namely, where the partial order is built on shared arguments? Extending our examples, we could build a theory of consisting of recommendations for women aged 55-75; we could ensure that such a theory corresponds to a section, and, if they can consistently be put together, a global section.

Thirdly, even if the formal models provide us with deeper understanding of the phenomena of textual disagreements and contradictions, it is not obvious that a sheaf-based model would improve any standard metrics of computational text understanding, e.g. allow us to more accurately compute the entailment relations. Thus, we will be creating computational models and measuring their performance.

\subsection{Does context and scalar implicature transform disagreements into contradictions?}
 
One of the referees of this paper commented that 
"once you add context, there are more contradictions than there are disagreements", and observed that recommendations
are often quite precise about e.g. about recommended daily allowance, adequate intake and maximums for various nutrients. In addition, "explicitly stating these three intervals means there are less opportunities for disagreements, and more opportunities for contradictions, between recommendations".  
Similarly, scalar implicature\footnote{\url{https://plato.stanford.edu/entries/implicature/}}, removes multiple interpretations of formally underspecified sentences; e.g. the sentence "I have three children" is consistent with having four children, but the implicature disallows the latter interpretation. And thus disagreements might in fact be contradictions given a sufficiently rich representations. This opens the question of how useful in practice is the distinction between disagreements and contradictions, in both ordinary and expert contexts.

We agree that question needs to be investigated. Our intuition is that typical conversations tend to fall somewhere between complete agreement and complete disagreement, and thus a partial common model constructed by the interlocutors is negotiated. Also in question answering, answers are often synthesized from multiple sources. For example, IBM Watson playing the Jeopardy! game used multitudes of snippets gathered from the internet prior to the game (\cite{chu2012textual}, \cite{schlaefer2011statistical}) .

We feel the approach proposed here captures aspects of these processes, and it is also applicable in other situations requiring building an interpretation based on information coming from multiple sources.

\subsection{Can we compute causes of disagreement?} 

A related issue is finding the reasons for disagreement. For example, the two theories of frying bacon we mentioned in the Introduction differ in specified temperature and the number of cooking cooking sheets mentioned. But the two parameters are not independent: in this case the heat transfer is slower with two sheets, hence higher temperature. How would we go from identifying disagreement to understanding the reasons behind it?

\subsection{Computing the lattice of parameters.} 

How can we compute the lattice of parameters for the sentences in different documents? For some types, such as time periods, and other measurement units, there has been enough work to make such computations clearly feasible (e.g.
\cite{hobbs06time},\cite{welty2006reusable,dAquin2012WhereTP}; biomedical research has a strong subarea focusing on ontologies (e.g. \cite{ivanovic2014overview}). But what about spacial relations like "riding bicycle on the beach" vs. "riding bicycle on the sand on the beach". Perhaps general purpose ontologies could be used to provide the required contextual information; for example, "beach" can be related through an ontology to "sea coast"\footnote{\url{http://www.adampease.org/OP/}, \url{http://sigma.ontologyportal.org:8080/sigma/TreeView.jsp?lang=EnglishLanguage&flang=SUO-KIF&kb=SUMO&term=Seacoast} retrieved on Oct 25 2017}, making it possible to create a taxonomy of related terms. It is less clear how to make plausible inferences using such taxonomies.

\section{Conclusions}

The sheaf model we sketched in this paper achieved the objectives of (a) identifying the causes of inconsistencies; (b) measuring their strength; (c) and suggesting ways to reconciling disagreements, if possible. We also discussed interesting connections to other areas of inquiry, possible new avenues of research, as well as the limitations of this work. 
Although this is definitely one of the first applications of sheaves to semantics of natural language, and one of the few appearances of sheaves in AI, obviously we believe that this particular set of mathematical methods can be used in other contexts, beside the few mentioned in this paper. We hope that the detailed discussion of related work and open issues will help with finding such new applications.

As commented earlier, whether sheaves can give us better computational models is an open question, and, in our opinion, it is worth researching. \\

\noindent
\textbf{Acknowledgments:} We would like to thank the referees for their comments, and for suggesting connections to methods of resolving contradictions in knowledge bases and to scalar implicature. We managed to only partially discuss these issues in this paper.

\bibliographystyle{named}
\bibliography{coguidelines}

\end{document}